\documentclass{article}
\usepackage[accepted]{icml2025}
\pdfoutput=1  % ADDED ANDRE
\usepackage{microtype}
\usepackage{graphicx}
\usepackage{subfigure}
\usepackage{booktabs} % for professional tables

\usepackage{hyperref}

\usepackage{icml2025}

\usepackage{amsmath}
\usepackage{amssymb}
\usepackage{mathtools}
\usepackage{amsthm}

\usepackage[capitalize,noabbrev]{cleveref}

\theoremstyle{plain}
\newtheorem{theorem}{Theorem}[section]

\theoremstyle{definition}
\newtheorem{definition}[theorem]{Definition}

\theoremstyle{remark}

\usepackage[textsize=tiny]{todonotes}

\icmltitlerunning{Rotary Offset Features in Large Language Models}

\begin{document}

\twocolumn[
\icmltitle{Rotary Offset Features in Large Language Models}

% It is OKAY to include author information, even for blind
% submissions: the style file will automatically remove it for you
% unless you've provided the [accepted] option to the icml2025
% package.

% List of affiliations: The first argument should be a (short)
% identifier you will use later to specify author affiliations
% Academic affiliations should list Department, University, City, Region, Country
% Industry affiliations should list Company, City, Region, Country

% You can specify symbols, otherwise they are numbered in order.
% Ideally, you should not use this facility. Affiliations will be numbered
% in order of appearance and this is the preferred way.
\icmlsetsymbol{equal}{*}

\begin{icmlauthorlist}
\icmlauthor{André Jonasson}{inst}
\end{icmlauthorlist}

\icmlaffiliation{inst}{Annokvick, Stockholm, Sweden}

\icmlcorrespondingauthor{André Jonasson}{mail@andrejonasson.com}

% You may provide any keywords that you
% find helpful for describing your paper; these are used to populate
% the "keywords" metadata in the PDF but will not be shown in the document
\icmlkeywords{Language Models, Rotary Position Embeddings, Feature Analysis, Attention Mechanism, Outlier Features, Attention Sinks}

\vskip 0.3in
]

% this must go after the closing bracket ] following \twocolumn[ ...

% This command actually creates the footnote in the first column
% listing the affiliations and the copyright notice.
% The command takes one argument, which is text to display at the start of the footnote.
% The \icmlEqualContribution command is standard text for equal contribution.
% Remove it (just {}) if you do not need this facility.

% \printAffiliations{}
% \printNotice{}
\printAffiliationsAndNotice{}

\begin{abstract}
Transformer-based Large Language Models (LLMs) rely on positional encodings to provide sequence position information to their attention mechanism. Rotary Positional Encodings (RoPE), which encode relative position by rotating queries and keys, have become widely used in modern LLMs. We study the features and patterns that emerge in queries and keys when using rotary embeddings and introduce the concept of rotary offset features. Our analysis reveals that these features, which frequently exhibit large activations and are often interpreted as outliers, arise consistently across layers, attention heads, and model architectures. We derive bounds predicting which rotary frequencies give rise to rotary offset features and the minimum angle between the query-key pairs for these features. We verify our predictions empirically across models of different sizes and architectures.
\end{abstract}

\section{Introduction}
\begin{figure}[t!]
\centering
\includegraphics[width=1\columnwidth]{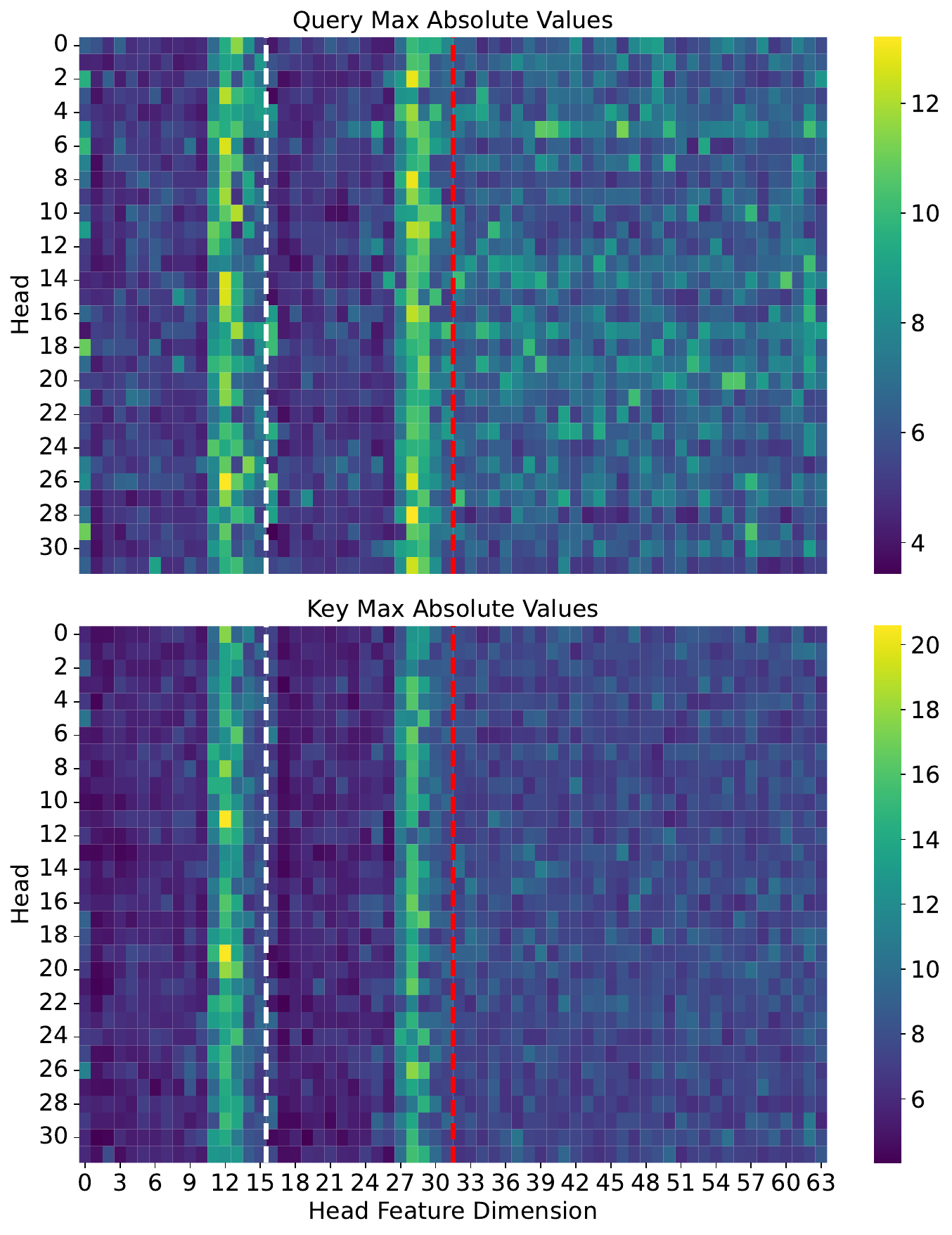}
\caption{Maximum query and key absolute values across all layers in Phi-1. The red dashed line separates rotary (left) from non-rotary (right) features, we further separate the first dimensions of the rotary pairs (left) from the second dimensions (right) with a white dashed line. The two bands of large absolute values are located at the pair of rotary dimensions that are rotated by the same low frequencies in a sliced rotary implementation layout. See Appendix Figure \ref{fig:query_key_stats_appendix} for mean magnitudes.}
\label{fig:query_key_stats}
\end{figure}

Rotary positional embeddings \citep{su2024roformer} have become a standard component in modern language models, yet their learned feature patterns remain poorly understood and are an active area of research \citep{barbero2024round}. While they have demonstrated strong performance across a range of model architectures and sizes, the underlying mechanisms by which they encode and utilize positional information are not well characterized.

In this paper, we analyze the internal representations of models using rotary positional embeddings (RoPE) to better understand how positional information is encoded and utilized. Through empirical analysis across multiple model architectures, we discover that queries and keys exhibit consistently large magnitudes at specific pairs of rotary feature dimensions (see Figure \ref{fig:query_key_stats}). We develop bounds that characterize these pairs of large-magnitude features, which we term rotary offset features, and demonstrate how they influence attention patterns.

Feature magnitudes are particularly relevant in the context of model quantization. Work by \citep{dettmers2022gpt3} has shown that activation outlier features can significantly impact model performance during quantization. While their analysis focused primarily on activation outliers in models with additive positional encodings, similar features exist for networks with rotary positional embeddings. The high-norm features we identify in rotary embeddings could affect the quantization errors for activation quantized models. This is because high-norm features tend to dominate the representation space, leading to larger quantization errors when the numerical precision is reduced. Furthermore, since keys contain these features, quantized key caches may require higher precision or more complex quantization algorithms \citep{liu2024kivi} to maintain model accuracy.

In the analysis we choose to focus solely on rotary features. We analyze the features and patterns that emerge in queries and keys when using rotary embeddings and make the following contributions:
\begin{itemize}
\item We find features with large magnitudes appear in pairs of feature dimensions at certain rotary frequencies, which we call rotary offset features, across layers and heads within networks and that they exist across language models of different sizes and architectures.
\item We show how a simple upper bound on the rotary frequency can be used to predict which frequencies are likely to contain rotary offset features. This upper bound depends only on context length and rotary frequency.
\item We derive a novel and simple lower bound for predicting the minimum angle the query-key pair has for the rotary offset features. This lower bound also only depends on context length and rotary frequency.
\item We present other properties and analysis of rotary features, such as a mechanism by which attention sinks are attended to in heads with rotary offset features and how context length extension affects low frequency rotary features in Llama-3 (Section \ref{sec:llama_3_series_and_context_length_extension}).
\end{itemize}

\section{Background}
\subsection{Attention Mechanism}
The attention mechanism enables language models to dynamically focus on different parts of the input sequence \citep{vaswani2017attention}. For sequence length $n$, attention operates on queries $Q$, keys $K$, and values $V$ in $\mathbb{R}^{n \times d}$, computing:

\begin{equation}
    \text{Attention}(Q,K,V) = \text{softmax}\left(\frac{QK^T}{\sqrt{d}} + M\right)V
\end{equation}

where $M$ is a causal mask ensuring each position only attends to previous tokens:
\begin{equation}
    M_{ij} = \begin{cases}
        0 & \text{if } i \geq j \\
        -\infty & \text{otherwise}
    \end{cases}
\end{equation}

Multi-head attention splits this computation into $h$ parallel heads, each with dimension $d_h = d/h$:

\begin{equation}
    \text{MultiHead}(Q,K,V) = \text{Concat}(\text{head}_1,\ldots,\text{head}_h)W^O
\end{equation}

where each $\text{head}_i$ applies the attention mechanism to learned projections of $Q$, $K$, and $V$. This structure allows the model to jointly attend to information from different representation subspaces \citep{vaswani2017attention}.

\subsection{Rotary Positional Embeddings}

Rotary positional embeddings \citep{su2024roformer} apply position-dependent rotations to queries and keys, by pairing features in the
embedding and applying the same two-dimensional rotation to each pair. For position $m$ and feature dimension $i$, the rotation matrix $R_{\Theta}(m,i)$ is defined as:
\begin{equation}
    R_{\Theta}(m, i) = \begin{bmatrix}
        \cos(m\theta_i) & -\sin(m\theta_i) \\
        \sin(m\theta_i) & \cos(m\theta_i)
    \end{bmatrix}
\end{equation}

where $\theta_i = c^{-2i/d}$ controls the frequency for dimension $i$, $i$ ranges from $0$ to $\frac{r}{2}$ where $r$ is the number of rotary feature dimensions, and $c$ is a hyperparameter. Note that the step size $\theta_i$ and frequency decrease as $i$ increases. This means that features with higher index have lower frequencies. For query vector $\mathbf{q}$ and key vector $\mathbf{k}$ at position $m$, rotary embeddings transform pairs of features\footnote{Note that there are several possible layouts for RoPE.}:
\begin{equation}
    \mathbf{k}_{m,\langle i \rangle} = \begin{bmatrix} k_{m,i} \\ k_{m,i+\frac{r}{2}} \end{bmatrix}
\end{equation}

\begin{equation}
    \text{RoPE}(\mathbf{k}_m)_i = R_{\Theta}(m, i) \mathbf{k}_{m,\langle i \rangle}
\end{equation}
and similarly for $\mathbf{q}$, where $\langle i \rangle$ denotes the $i$-th rotary pair. This rotation preserves the inner product structure while encoding relative positions. For queries and keys at positions $m$ and $n$ respectively, the inner product between their rotated features preserves relative position information:

\begin{align}
    \text{RoPE}(\mathbf{q}_m, m)_{i}^T &\text{RoPE}(\mathbf{k}_n, n)_{i} \\&= \mathbf{q}_{m,\langle i \rangle}^T R_{\Theta}(m, i)^T R_{\Theta}(n, i) \mathbf{k}_{n,\langle i \rangle} \\
    &=(R_{\Theta}(m-n, i)\mathbf{q}_{m,\langle i \rangle})^T \mathbf{k}_{n,\langle i \rangle}
\end{align}

This shows that the inner product depends only on the relative position $(m-n)$ between the query and key, enabling the model to attend based on token distances. For networks with partial rotary embeddings, only a subset of the query and key feature dimensions receive rotations, with $r < d$.

The rotary embeddings enter into the attention mechanism by modifying the query-key dot products that determine attention weights. For a query $\mathbf{q}_m$ at position $m$ and key $\mathbf{k}_n$ at position $n$, the dot product becomes:

\begin{align}
    \mathbf{q}_m^T\mathbf{k}_n &= \sum_{i=1}^{r/2} (R_{\Theta}(m-n, i)\mathbf{q}_{m,\langle i \rangle})^T \mathbf{k}_{n,\langle i \rangle}\\
    &+\sum_{i=r+1}^{d_h} q_{m,i}k_{n,i}
\end{align}

where the first sum is over rotary feature pairs and the second sum is over non-rotary features. The rotary features enable position-dependent attention through the rotation matrices $R_{\theta}(n-m, i)$, while non-rotary features contribute position-independent semantic matching. After scaling by $1/\sqrt{d_h}$ and applying the causal mask, these modified dot products enter the softmax to produce attention weights:

\begin{equation}
    \alpha_{mn} = \frac{\exp(\mathbf{q}_m^T\mathbf{k}_n/\sqrt{d_h} + M_{mn})}{\sum_{j=1}^n \exp(\mathbf{q}_m^T\mathbf{k}_j/\sqrt{d_h} + M_{mj})}
\end{equation}

The attention weights $\alpha_{mn}$ then combine the value vectors to produce the output. This mechanism allows the model to learn both position-aware and position-independent attention patterns through the selective use of rotary and non-rotary features.

\section{Methodology}
\subsection{Decomposing positional attention patterns}
Instead of working with the $n$ query and key vectors and $n \times n$ dot products to understand the effect of the rotary features on the attention score, we simplify our analysis of the effect of the rotary features by working with the mean query and key vectors.
\begin{figure}[t]
    \centering
    \includegraphics[width=\columnwidth]{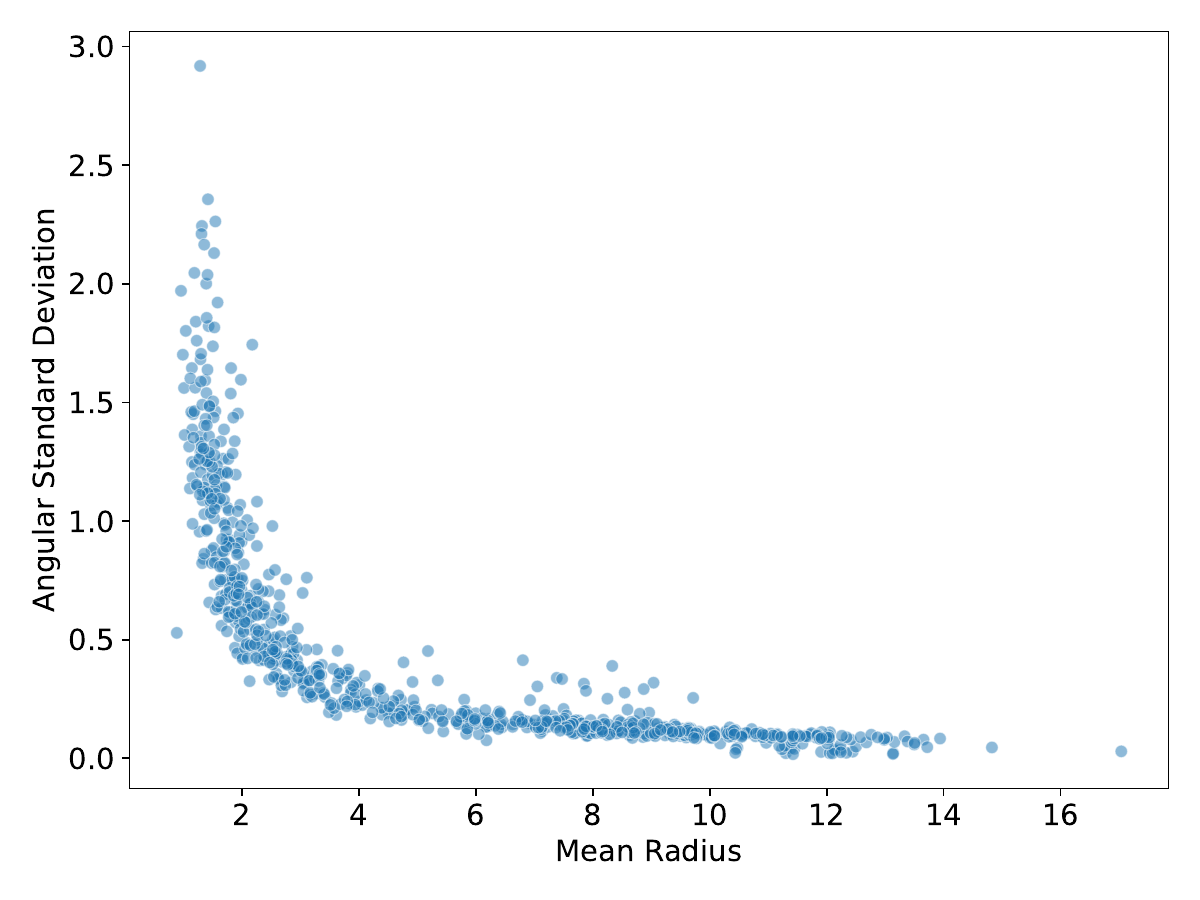}
    \caption{Circular angular standard deviation \citep{Mardia1999a} vs radius for feature 12, other features follow a similar curve, where angular standard deviation decreases rapidly as radius increases.}
    \label{fig:angular_std_vs_radius_f12}
\end{figure}
For each rotary feature pair $i$, we can compute the dot product between the mean query and mean key vectors. We can then rotate the queries to get a sense for the evolution of the dot product contribution of the rotary feature pair with distance. Let $\bar{\mathbf{q}}_{\langle i \rangle}$ be the mean query vector and $\bar{\mathbf{k}}_{\langle i \rangle}$ be the mean key vector across all positions for rotary pair $\langle i \rangle$. We adopt the view that the query is incrementing its position $p$ relative to the key and the key is stationary, which gives us the dot product per rotary feature pair:
\begin{align}
    d_i(p) &= (R_{\Theta}(p, i)\bar{\mathbf{q}}_{\langle i \rangle})^T  \bar{\mathbf{k}}_{\langle i \rangle} \\
    &= \|\bar{\mathbf{q}}_{\langle i \rangle}\| \|\bar{\mathbf{k}}_{\langle i \rangle}\| \cos(\phi_i + \theta_i p)
\end{align}
where $\phi_i\in(0, 2\pi]$ is the initial counterclockwise angle between $\bar{\mathbf{q}}_{\langle i \rangle}$ and $\bar{\mathbf{k}}_{\langle i \rangle}$. This gives us an $\frac{r}{2}$-dimensional vector function $\mathbf{d}$ where each index $i$ contains the dot product between the counter-clockwise rotated mean query and the mean key vector at a positional distance $p$. In the rest of this paper, when we mention "radius", we will be speaking about the rotary feature pair's mean vector's magnitude $\|\bar{\mathbf{x}}_{\langle i \rangle}\|$.

We find that rotary feature pairs with larger mean radius tend to have more stable angles with respect to the origin (Figure \ref{fig:angular_std_vs_radius_f12}) so we expect this simplification to preserve the key patterns as the noisier a feature's angle is the less likely it is to contribute significantly to the dot product. However, some rotary feature vectors deviate significantly from the mean, often serving as attention sinks, and this simplification will not capture these effects.

\begin{figure}[t]
    \centering
    \includegraphics[width=1\columnwidth]{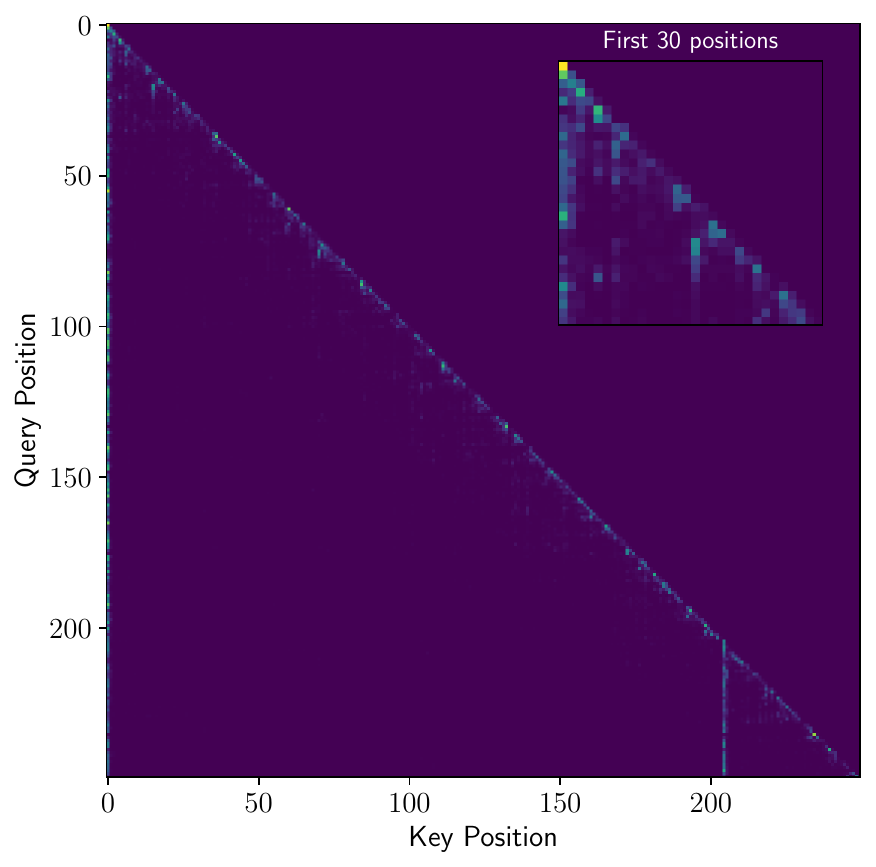}
    \caption{Attention pattern for a single head showing characteristic sub-diagonal structure and two attention sinks (position 0 and 204).}
    \label{fig:main_attention}
\end{figure}

\begin{figure}[t]
    \centering
    \includegraphics[width=\columnwidth]{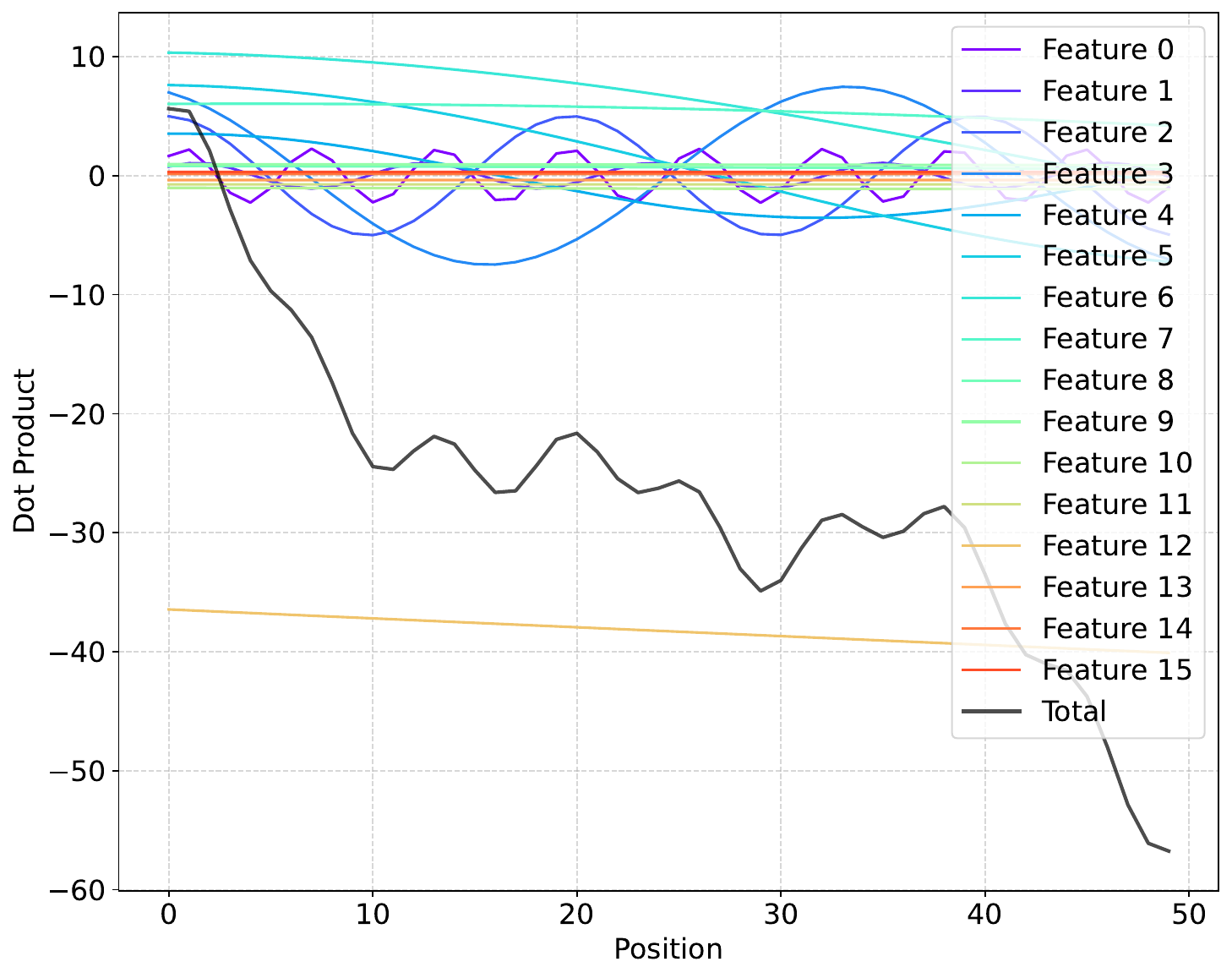}
    \caption{Each rotary feature's dot product component of $\mathbf{d}(p)$ for a range of positions $p$ and the total sum of the components showing the unnormalized attention score as distance increases. Feature 12 starts at about $-35$ at $p=0$. The positional attention pattern focuses on the first three tokens. The reconstructed positional attention pattern can be seen in Figure \ref{fig:decomposed_attention_pattern}.}
    \label{fig:decomposed_rotations}
\end{figure}

We calculate $\mathbf{d}(p)$ for a range of distances and plot each component $i$ as a function of $p$ and their sum in Figure \ref{fig:decomposed_rotations} and after scaling the sum $D(p)=\Sigma_i d_i(p)$ by $1/\sqrt{d_h}$ and applying the causal mask, we arrive at the resulting positional attention pattern. Comparing Figure \ref{fig:decomposed_rotations} with Figure \ref{fig:main_attention}, we can see that $D(p)$ recovers a focus on the first few tokens along the diagonal and quickly decays.

From this point on, we will use "feature" or "rotary feature" to refer to rotary feature pairs, since we will solely discuss rotary feature pairs.

%See Appendix for more examples.

\subsection{Rotary Offset Features and Bounds}
What we find is that most of the rotary features with large radius are concentrated in rotary frequencies which never complete a full period during the context length. Furthermore, the ones that turn into large-radius features within low frequencies tend to have a U-shaped contribution to $D(p)$, starting at a greater point than in the middle and often ending at a lesser point than they started. We call these features rotary offset features, and define them as:
\begin{definition}
    A rotary feature $i$ is a \textbf{rotary offset feature} if the dot product between mean query and mean key vectors are less than the starting dot product, $d_i(p) < d_i(0)$, for all positions less than or equal to the the context length, $p\leq p_{max}$.
\end{definition}

\begin{figure}[t]
    \centering
    \begin{tikzpicture}[scale=2]
        % Draw the unit circle
        \draw (0,0) circle (1);

        % Draw the axes
        \draw[->] (-1.2,0) -- (1.2,0) node[right] {$x$};
        \draw[->] (0,-1.2) -- (0,1.2) node[above] {$y$};

        % Draw two vectors at an obtuse angle
        \draw[->, line width=1.1pt] (0,0) -- (0.866,0.5) node[above right] {$\vec{q_0}$};
        \draw[->, line width=1.1pt, dashed] (0,0) -- (-0.866,0.5) node[above left] {$\vec{q_1}$};
        \draw[->, line width=1.1pt] (0,0) -- (0.0,-1.0) node[below left] {$\vec{k}$};

        % Draw a simple arc showing the obtuse angle
        \draw [dashed,domain=30:151] plot ({0.3*cos(\x)}, {0.3*sin(\x)});
        \draw [domain=30:270] plot ({0.15*cos(\x)}, {0.15*sin(\x)});
        \node at (0.1,0.5) {$2\phi-2\pi$};
        \node at (-0.2,-0.15) {$\phi$};
        \draw [dashed,domain=155:270] plot ({0.5*cos(\x)}, {0.5*sin(\x)}) node[below left] {$2\pi-\phi$};
    \end{tikzpicture}
    \caption{Query $\vec{q_0}$ and key $\vec{k}$ at an angle $\phi>\pi$. Using the relationship $cos\phi=cos(2\pi-\phi)$, $\phi-(2\pi-\phi)$ is the the maximum rotation of $q_0$ for it to return to same dot product, $\vec{q_1}^T\vec{k}=\vec{q_0}^T\vec{k}$, without exceeding the original dot product along the rotation.}
    \label{fig:obtuse_angle}
\end{figure}

Using this definition, we can reason around the properties of rotary offset features and then test empirically whether the properties hold. Firstly, for the rotary offset feature to not return to the same value as at zero distance, the period of the rotation $T=2\pi/\theta$ must be greater than the context length $p_{max}$, giving us the upper bound on the rotary frequency:
 \begin{equation}
    \theta < \frac{2\pi}{p_{max}}
\end{equation}

Secondly, we know that the initial angle between query and key needs to be $\phi>\pi$, since $\cos\phi$ is increasing for $[\pi, 2\pi]$ the rotary offset feature would be increasing above its initial value $d_i(0)$ for some $p\leq p_{max}$. To arrive at a tighter lower bound for the initial angle, we need to use that $cos(2\pi-\phi)=cos(\phi)$, and the angular change $\phi-(2\pi-\phi)$ gives us the angle at which the rotary offset feature returns to the same value as at zero distance (see Figure \ref{fig:obtuse_angle}). This gives us the inequality $p_{max} \theta \leq 2\phi-2\pi$, where the left hand side is the total rotation at the context length and the right hand side is the maximum change in angle to return to the value at zero distance. Rearranging this, we arrive at a lower bound for the query-key angle:
\begin{equation}
    \phi > \pi + \frac{p_{\text{max}}\theta}{2}
\end{equation}

These bounds only apply to rotary offset features, which tend to be the cause of the majority of rotary features with large radius.

\section{Results}
\subsection{Model Selection}
We focus our analysis on Phi-1 \citep{gunasekar2023textbooksneed}, due to its low amount of rotary features which simplifies communication of results, and to a lesser extent DeepSeek-V2-Lite \citep{liu2024deepseek}, which has significant architectural differences to Phi-1, including multiheaded latent attention and a single rotary key for all rotary query heads, and Llama-3-8B/70B \citep{dubey2024llama}. The partial rotary embeddings should let us separate semantic from positional features and we should be able to observe the effects of the rotary positional embeddings more clearly.

\begin{table}[h]
\centering
\begin{tabular}{@{}lccccc@{}}
\hline
\textbf{Model} & \textbf{d\textsubscript{h}} & \textbf{r} & \textbf{p\textsubscript{max}} & \textbf{extended}\\
\hline
Phi-1 & 64 & 32 & 2048 & No\\
Llama-3-8B & 128 & 128 & 4096 & No\\
Llama-3-70B & 128 & 128 & 8192 & No\\
DeepSeek-V2-Lite & 192 & 64 & 163840 & Yes\\
\hline
\end{tabular}
\caption{Model architecture details showing head dimension ($d_h$), rotary dimension ($r$), context length ($p_{max}$), and whether the linear projections for keys and queries have bias.}
\label{tab:model_details}
\end{table}
%present inv freq $c$ in table

Throughout the paper, we use zero-based indexing for layers, heads, and features and will ignore non-rotary features. For all statistics and plots involving queries and keys, we analyze the data before applying rotary transformations, unless otherwise noted. We use rotary first with a sliced rotary implementation layout (as in most \texttt{transformers} models' implementations \citep{wolf-etal-2020-transformers}), we reorder DeepSeek-V2-Lite's activations\footnote{DeepSeek-V2-Lite has rotary last interleaved layout.} to follow this layout as well. For this layout, the rotary features' frequencies decrease as the feature index increases.

We use the configured max position embeddings of models, $p_{max}$, as a proxy for trained context length. This assumes that the models were trained with these sequence lengths.

\subsection{Single Head Analysis}
We begin our analysis with a single head\footnote{Layer 2 head 4 (zero-indexed) of Phi-1.} in Phi-1. We will later show that the patterns are in many other heads and layers, followed by an analysis showing the patterns exist across different architectures and sizes of language models.

\paragraph{Cyclic Rotary Features}
Various positional attention patterns can be constructed using the rotary features that complete several cycles during the context length, as they form a combination of sinusoidal functions of various amplitudes and frequencies. For instance, in heads with banded attention patterns, as in the head we have chosen to study, seen in Figure \ref{fig:main_attention}, uses multiple low-to-mid frequency components (as seen in the decomposition in Figure \ref{fig:decomposed_rotations}) to build up a maximum attention score at distance zero and one and then quickly decays, forming a banded diagonal pattern.

\paragraph{Partial Cycle Rotary Features}
Some of the most consistent large-radius features across heads and layers (and as we'll see, across different models) are the features that go through exactly a full period or less of the RoPE rotations during the model's context length. These tend to be selected to ensure the attention score does not return to the same value as at early distances. See Figure \ref{fig:decomposed_rotations} for an example where feature 12 starts at a large negative value.

\begin{figure}[t]
    \centering
    \includegraphics[width=0.7\columnwidth]{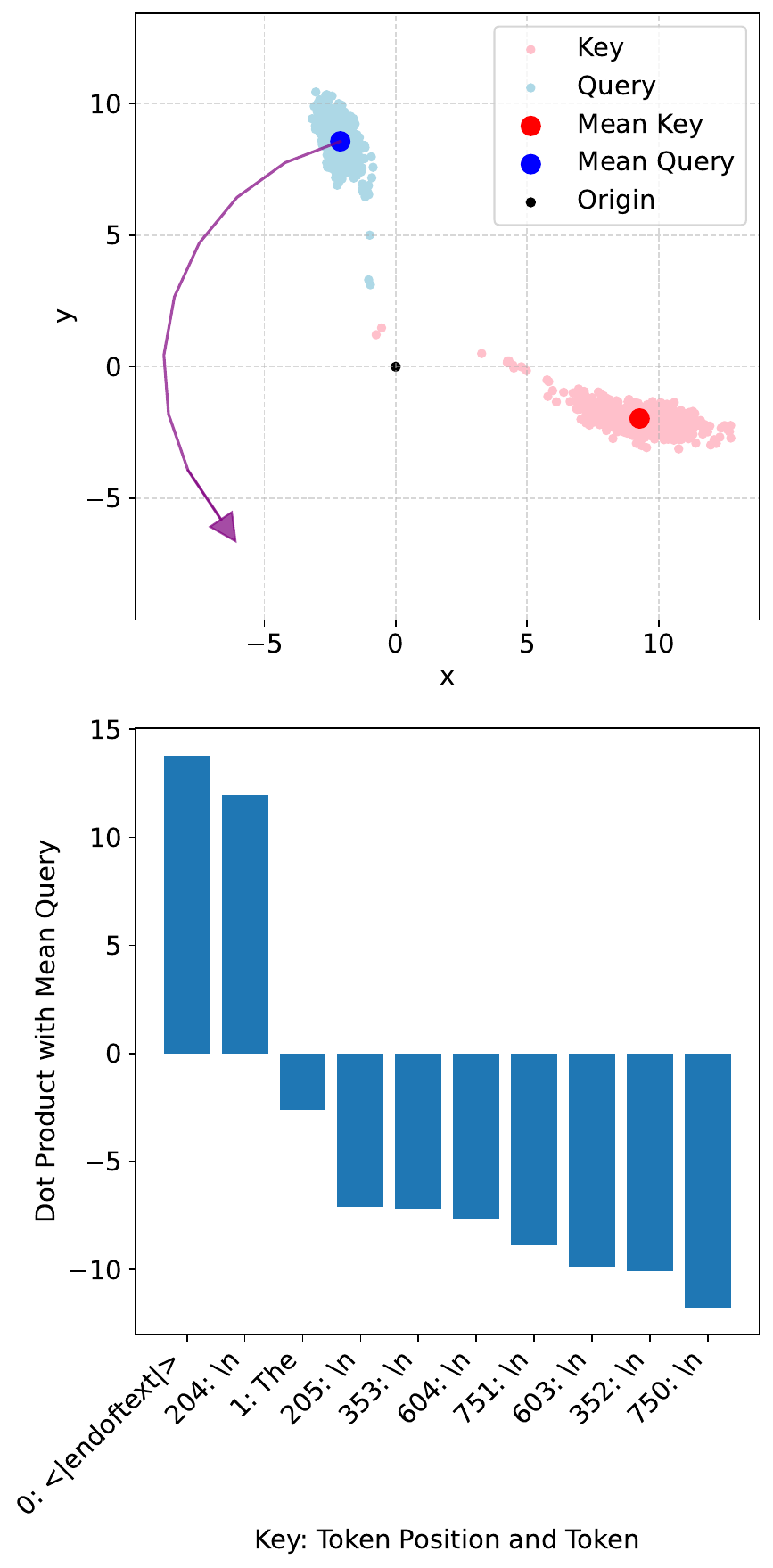}
    \caption{Feature 12's keys and queries are shown, with their respective means. The curved line shows the trajectory of the mean query rotated with frequency $\theta_{12}$ across the network's context length, $p_{max}$. The query and key means maintain angles that yield a consistently negative dot product. The two pink dots just above the origin are the attention sink keys, which align with the query angle. In the bottom plot, we show that these are the only two tokens (out of over 800) that produce positive dot products, causing them to receive a large proportion of the attention weight.}
    \label{fig:feature_12}
\end{figure}

\begin{figure*}[!t]
    \centering
    \includegraphics[width=15cm]{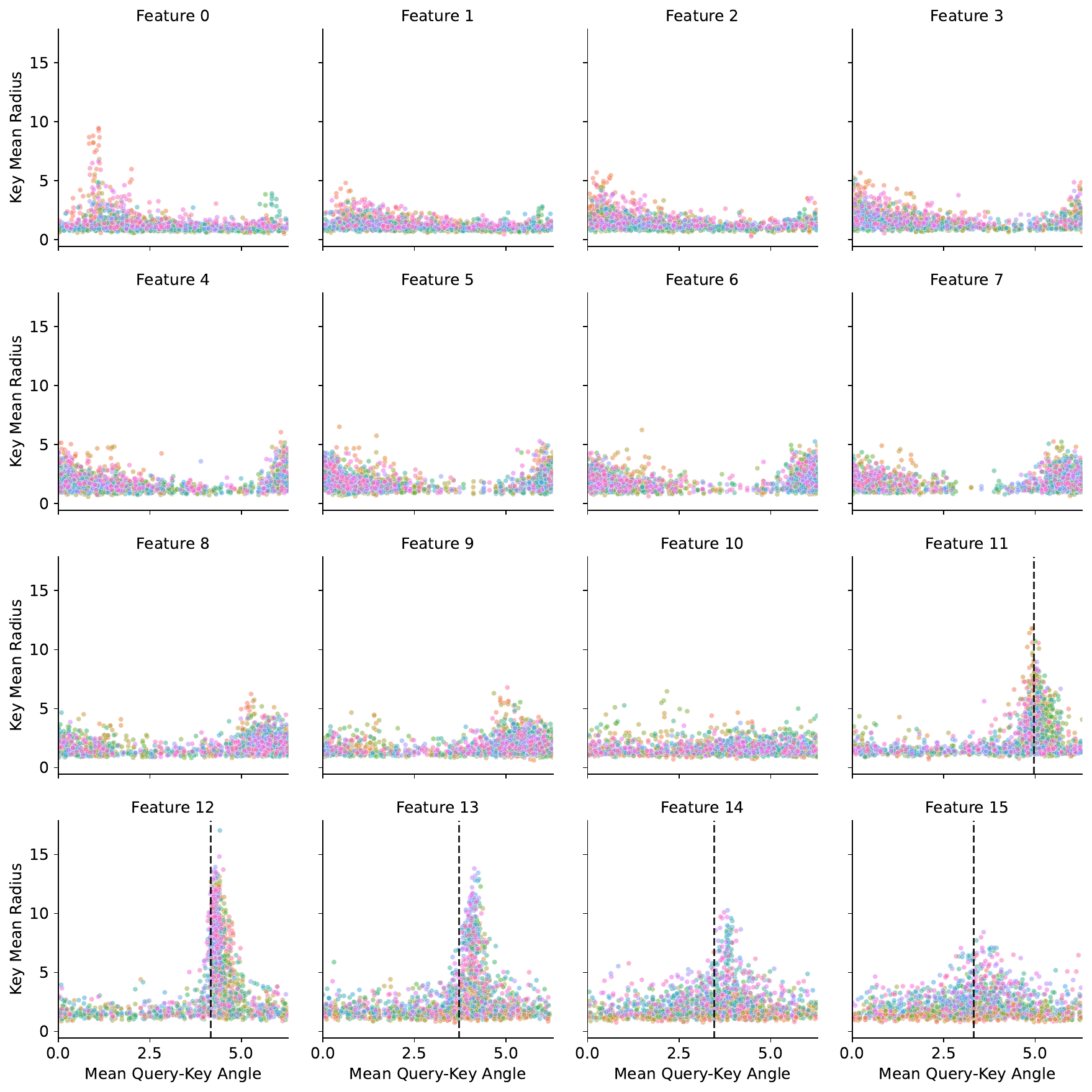}
    \caption{Mean radius of rotary key features plotted against the angle between mean query and mean key for each layer (colored) and head of Phi-1 (768 datapoints for each feature). Black dashed line for feature 11-15 are the lower bounds for the angle $\phi_i$ for each feature to serve as a rotary offset feature. Any feature without a dashed line are outside the bounds.}
    \label{fig:angular_diff_vs_radius}
\end{figure*}

Rotary offset features are the ones that create the bands of large magnitudes in the query and key plots in Figure \ref{fig:query_key_stats}. In Phi-1, these bands are formed around feature 12. For the head we have selected for analysis, feature 12 is also the feature with greatest radius of the head's rotary features. In Figure \ref{fig:feature_12}, we see that the feature has a large radius in both query and key and maintains consistent angles with respect to the origin (with some exceptions discussed later). The angle between the query and key vectors for feature 12 tend to form an angle $\phi_{12}$ which would result in a significant negative dot product (see Figure \ref{fig:feature_12_xy_coords_wiki} for the other 15 features). We also show the full rotation throughout the context length, which covers less than half a full period. This head is not unique and the angle between query and key are surprisingly similar for features with large radius with index 12 across different heads and layers (see Figure \ref{fig:angular_diff_vs_radius}).

\begin{figure}[t]
    \centering
    \includegraphics[width=\columnwidth]{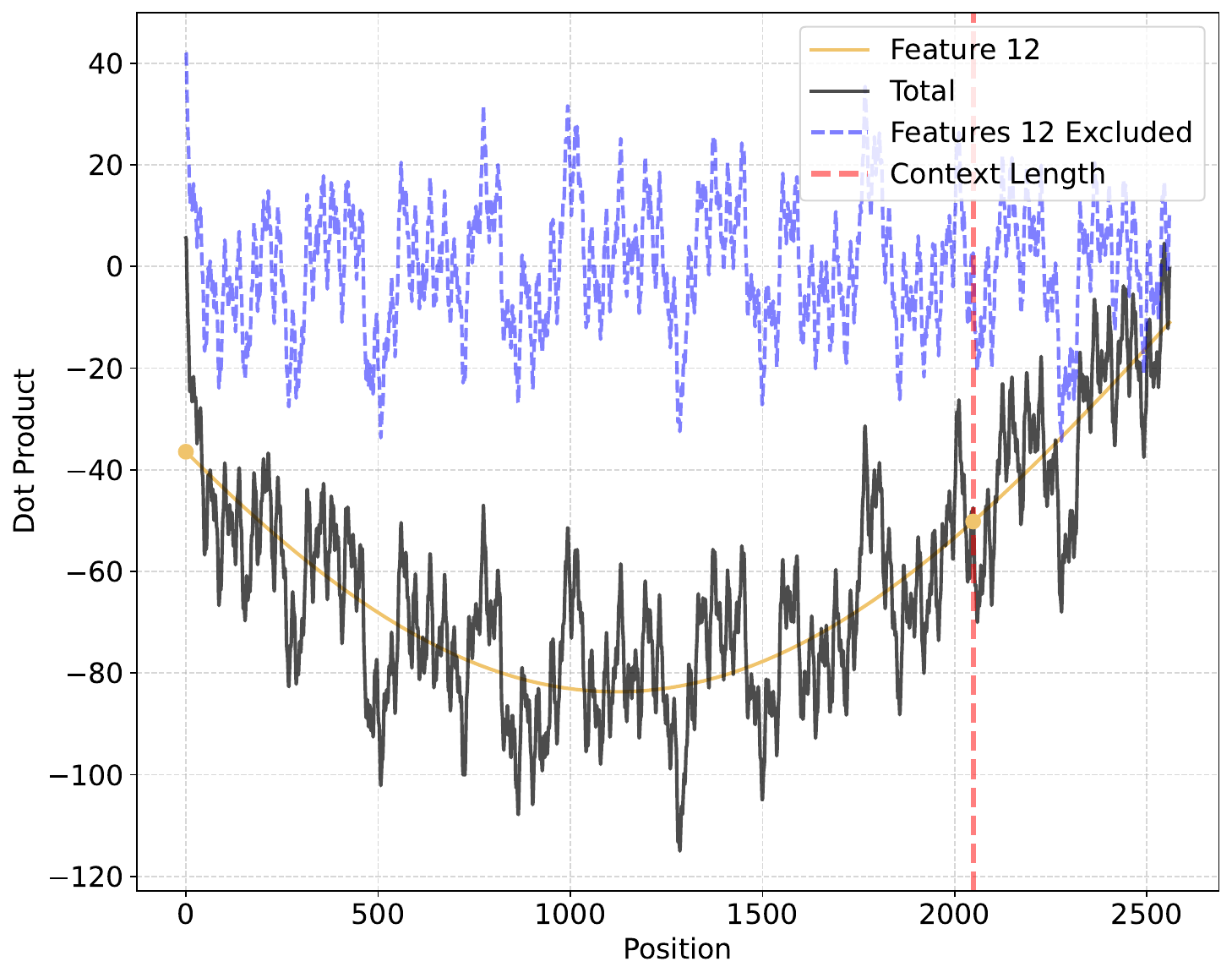}
    \caption{$D(p)$ for positions 0-2560 (512 past the context length of Phi-1) with the addition of a line for what happens when we exclude feature 12 and feature 12 without other features. Feature 12 never completes a cycle. Figure \ref{fig:feature_12} top plot shows the corresponding rotation during the context length.}
    \label{fig:beyond_max}
\end{figure}

To better understand the feature's contribution to the dot product $D(p)$, we study what happens when we exclude this feature from $D(p)$. In Figure \ref{fig:beyond_max}, we see that at short distances ($p\lesssim 50$) excluding feature 12, just shifts the curve up, which since softmax is invariant to adding a constant to all inputs, $softmax(\mathbf{x}+c) = softmax(\mathbf{x})$, would result in the same attention pattern. However, at longer distances there is a large difference in $D(p)$, at many points where the other features would add up to being close to values at the start, feature 12 has shifted them down, making their contribution to the attention score negligible. At the end, $d_{12}(p_{max})$ is closing in on its value at the start, but the other features, $d_{0:15\backslash12}$, do not peak at the same time as at the start which results in a relatively much lower attention score. In other words, without feature 12, the attention pattern would look very different at longer distances and would at certain intervals attend tokens that are distant.

\paragraph{Attention sinks} In Figure \ref{fig:feature_12}, we also see that Feature 12 plays a role in which tokens are selected as attention sinks \citep{xiao2024efficientstreaminglanguagemodels} in this head, with
\texttt{<end\_of\_text>} and the first \texttt{\textbackslash n} being selected. The head makes use of the low frequency rotary feature pair and aligns the keys of the attention sinks to the queries' mean angle resulting in a positive dot product across positions, it then proceeds to put all other keys at angles from the queries mean angle which result in consistently smaller dot product for the other keys, leading to a large amount of attention on the attention sinks leading to the typical vertical attention pattern, as seen at position 0 and 204 in Figure \ref{fig:main_attention}.

\subsection{Model Analysis}
\subsubsection{Phi-1}
\paragraph{General observations}
We observe that many of the relationships in the head we just analyzed are present across other layers and heads in Figure \ref{fig:angular_diff_vs_radius} (and across data samples \ref{fig:angular_diff_vs_mean_radius_key_code}). Large-radius features tend to be at certain angles $\phi_i$. It can be observed for Phi-1 that most of them are at low frequencies, with the exception of feature 0, which is a cyclic rotary feature. For Feature 0 the largest radii heads are located around $\phi=1 rad$ which is equal to the angular step per position, $\theta_0=1$, so the query and key will tend to maximise their dot product at a distance of one, suggesting these are  features that select for the previous token.

\paragraph{Rotary offset features cutoff} For Phi-1's feature 10, we calculate an upper bound of $p_{max}\theta_{10}=6.38$, which exceeds $2\pi$, this indicates that feature 10 and any higher frequency rotary features would be unsuitable as rotary offset features. We observe a sharp increase in maximum radii between features 10 and 11, aligning with our predictions about which features can effectively serve as rotary offset features.

\paragraph{Initial query-key angle of rotary offset features} We visualize the lower bounds in Figure \ref{fig:angular_diff_vs_radius}. By inspection, the model seems to agree with these bounds and the majority of large-radius features maintain angles above or very close to the lower bound for each rotary frequency.

\begin{table*}[t]
    \centering
    \begin{tabular}{lrrrrrrr}
        \toprule
        Model & Features & \%ROF & Mean LB & Min. Radius & Positives & UB Recall & LB / LB-0.1 Recall \\
        \midrule
            Phi-1 & 12288 & 31\% & 3.93 & 6.0 & 634 & 0.97 & 0.88 / 0.94 \\
             & & & & 9.0 & 252 & 0.99 & 0.94 / 0.98 \\
             & & & & 12.0 & 46 & 1.00 & 1.00 / 1.00 \\
             Llama-3-8B & 65536 & 45\% & 3.72 & 6.0 & 1888 & 0.90 & 0.63 / 0.78 \\
             & & & & 9.0 & 812 & 1.00 & 0.82 / 0.94 \\
             & & & & 12.0 & 344 & 1.00 & 0.84 / 0.99 \\
             Llama-3-70B & 327680 & 45\% & 3.72 & 6.0 & 14832 & 0.63 & 0.40 / 0.52 \\
 & & & & 9.0 & 5440 & 0.85 & 0.55 / 0.73 \\
 & & & & 12.0 & 2024 & 0.97 & 0.63 / 0.82 \\
            DeepSeek-V2-Lite$^\dagger$ & 13824 & 28\% & 4.26 & 6.0 & 640  & 0.92 & 0.66 / 0.84 \\
             & & & & 9.0 & 464 & 1.00 & 0.68 / 0.94 \\
             & & & & 12.0 & 352 & 1.00 & 0.74 / 0.97 \\
        \bottomrule
    \end{tabular}
    \caption{Metrics for various models for keys, showing the number of rotary features, percentage of rotary features that could be selected as rotary offset features (\%ROF), mean lower bound for the angles $\phi_i$ across all the rotary offset features (Mean LB), the minimum radius of rotary feature pairs for defining positives (Min. Radius), the number of features considered positives (Positives) for a particular minimum radius, recall of positives for the upper bound (UB Recall), and the recall of positives for the lower bound / relaxed lower bound (by 0.1 rad or $\sim$6$^{\circ}$)) (LB / LB-0.1 Recall) for the mean query-key angle. $\dagger$ For DeepSeek-V2-Lite, note that the rotary keys in these statistics are repeated 16 times since each rotary key is matched to 16 rotary queries.}
    \label{tab:metrics}
\end{table*}

\subsubsection{Metrics across models}
DeepSeek-V2-Lite and Llama-3-8B/70B also have rotary offset features, similarly to Phi-1, despite their different architectures and scales. Table \ref{tab:metrics} presents the recall metrics for both upper and lower bounds across our analyzed models. For the upper bound, we find that features with radius greater than radius 6.0 are within the bounds over 90\% of the time in Phi-1, DeepSeek-V2-Lite and Llama-3-8B. The lower bound, despite covering less than 40\% of possible angles between 28-45\% of all features, depending on the model, successfully captures the majority of large-radius features across Phi-1, DeepSeek-V2-Lite and Llama-3-8B. For Llama-3-70B, we see lower recall for radius 6.0 than for the other models, however from radius 9.0 and greater our bounds show good recall, and we again see a clustering around the lower bound since there's almost 20\% of features with large radius just below the lower bound. Visual evidence in Figures \ref{fig:angular_diff_vs_radius} (and Figure \ref{fig:deepseek_angular_diff_vs_mean_radius} for DeepSeek) further supports our findings, showing clear clustering of large-radius features around our predicted lower bounds. Relaxing our lower bound by 0.1 rad or $\sim$6$^{\circ}$ the recall for the lower bound improves, indicating that a concentration of large-radius features are also just below the lower bound.

Our predictions about the behaviour of rotary offset features and the empirical results from our bounds suggest that one of the main drivers for large-magnitude features in low-frequency rotary features are rotary offset features. The pair of feature dimensions that make up the rotary offset features correspond to the two high-norm bands seen in Figure \ref{fig:query_key_stats}.

\section{Related Work}

\citep{chen2024hope, liu2024scaling} studies length generalization in RoPE and hypothesizes that low-frequency features are unable to generalize to lengths beyond the context length. They derive an upper bound for the frequency of these features which turns out to be the same as the upper bound for our rotary offset features. We independently discovered this upper bound during our research on large-radius rotary features and show how it can instead be applied to predict which rotary frequencies are likely to contain large-magnitude activations.

\citep{barbero2024round} performs an in-depth analysis of RoPE. They also note that high frequency rotary feature pairs are used to create previous token (sub-diagonal) heads and note heavy use of low-frequency components in Gemma.

\citep{jin2025massivevaluesselfattentionmodules} studies massive values in large language models and found independently and concurrently to us massive values in queries and keys, that they are present in models using RoPE, and that they are found in rotary feature pairs.

Recent work has identified attention sinks \citep{xiao2024efficientstreaminglanguagemodels} as a key mechanism in language models, where certain tokens (typically beginning of the sequence and semantically unimportant tokens) persistently receive high attention scores across many attention heads. Our analysis reveals the rotary offset features as one mechanism of how attention sinks are consistently selected for across positions. \citep{gu2024attentionsinkemergeslanguage} also notes that attention sinks emerge when the cosine between queries and keys is positive and large. We show how this works on an individual query-key level for the rotary offset features.

\citep{liu2024kivi} studies how to improve kv-cache quantization. They note that high-norm channels in keys are not present in the value vectors. They choose to employ asymmetric channel-wise quantization for keys, among other techniques, to reduce the impact of these high-norm channels on the model's performance after quantization.

\section{Conclusion}
We have presented a detailed analysis of rotary features in language models, revealing consistent patterns across architectures and scales. Our analysis reveals several key findings about how transformers utilize rotary positional embeddings:

First, we demonstrate that certain rotary feature pairs consistently emerge with large radii across different models and architectures. These features tend to have a U-shaped contribution to the attention score with relative distance between tokens. We call these rotary feature pairs rotary offset features.

Second, we derive bounds that predict at which rotary frequencies rotary offset features emerge and above what angle the corresponding rotary feature pairs in the queries and keys tend to be. This matches empirical observations of where large-radius rotary feature pairs appear. These bounds solely rely on context length and rotary frequency as input.

We hope the derived bounds, the insights and the analysis presented in this paper can help further the understanding of models using rotary positional embeddings, lead to improvements in the frequency selection of rotary features for context length extension and inform the development of improved quantization algorithms for language models using rotary embeddings.

\section*{Impact Statement}
This paper advances understanding of language model architectures. The insights may enable more efficient implementations through improved quantization, potentially reducing computational costs and environmental impact of large language model deployment.

\bibliography{rotary_feature_analysis}
\bibliographystyle{icml2025}

%%%%%%%%%%%%%%%%%%%%%%%%%%%%%%%%%%%%%%%%%%%%%%%%%%%%%%%%%%%%%%%%%%%%%%%%%%%%%%%
%%%%%%%%%%%%%%%%%%%%%%%%%%%%%%%%%%%%%%%%%%%%%%%%%%%%%%%%%%%%%%%%%%%%%%%%%%%%%%%
% APPENDIX
%%%%%%%%%%%%%%%%%%%%%%%%%%%%%%%%%%%%%%%%%%%%%%%%%%%%%%%%%%%%%%%%%%%%%%%%%%%%%%%
%%%%%%%%%%%%%%%%%%%%%%%%%%%%%%%%%%%%%%%%%%%%%%%%%%%%%%%%%%%%%%%%%%%%%%%%%%%%%%%
\newpage
\appendix
\onecolumn
\section{Additional Figures for main text}
\begin{figure}[H]
    \centering
    \includegraphics[width=0.8\columnwidth]{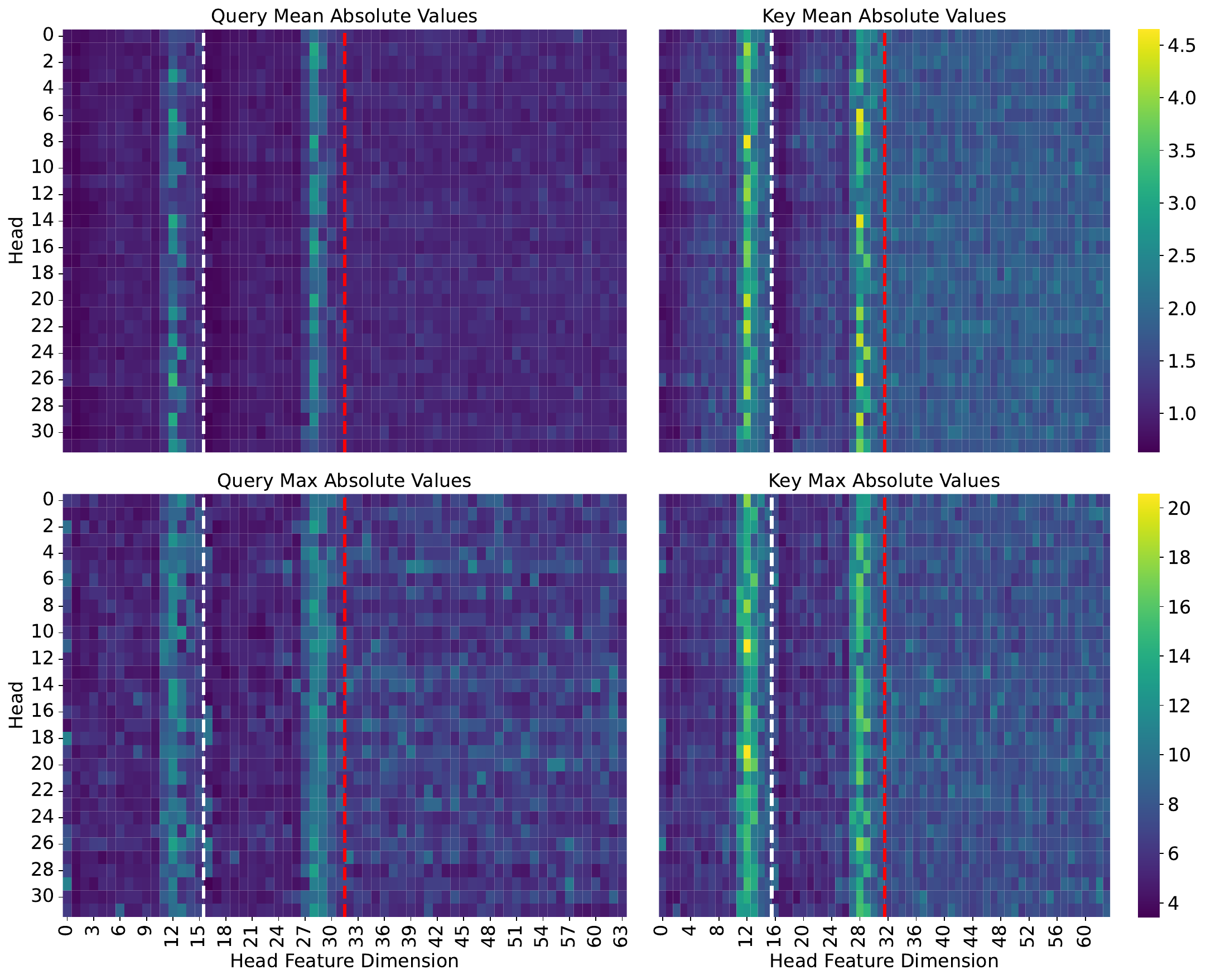}
    \caption{Query and key magnitudes prior to rotation in Phi-1. The red dashed line separates rotary (left) from non-rotary (right) features, we further separate the first dimensions of the rotary pairs (left) from the second dimensions (right) with a white dashed line. The bands of large absolute values are located at the pair of dimensions that are rotated by the same low frequency in a sliced rotary implementation layout.}
    \label{fig:query_key_stats_appendix}
\end{figure}

\begin{figure}[H]
    \centering
    \includegraphics[width=0.4\textwidth]{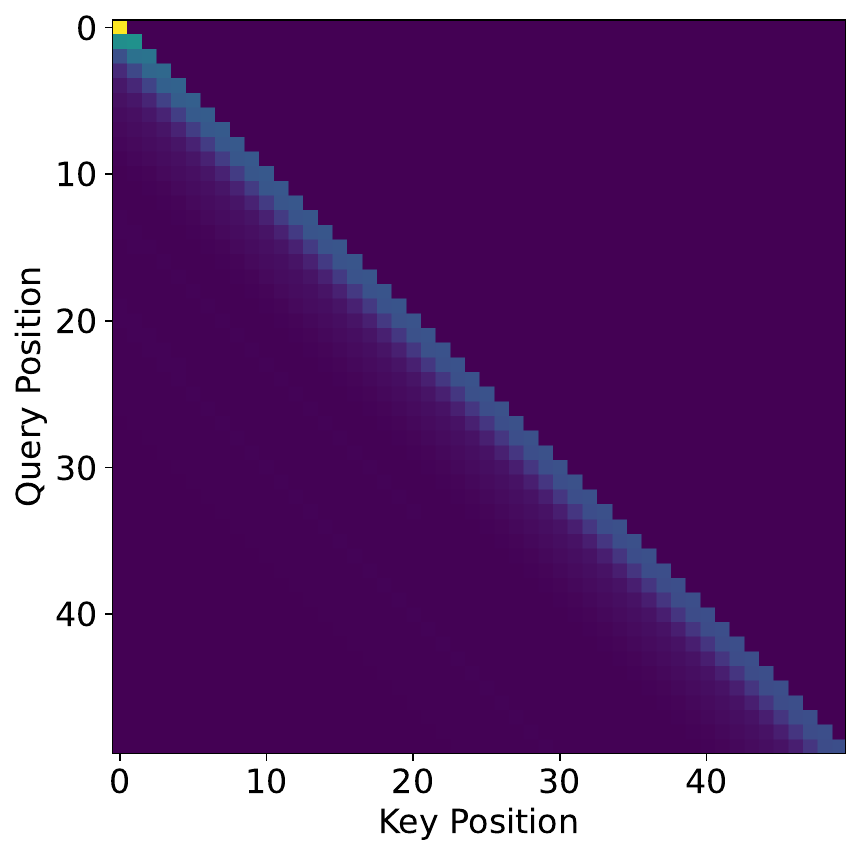}
    \caption{Attention pattern after applying scaling, masking and softmax to $\Sigma_i d_i(p)$ shown as total in Fig \ref{fig:decomposed_rotations}.}
    \label{fig:decomposed_attention_pattern}
\end{figure}

\begin{figure}[H]
    \centering
    \includegraphics[width=15cm]{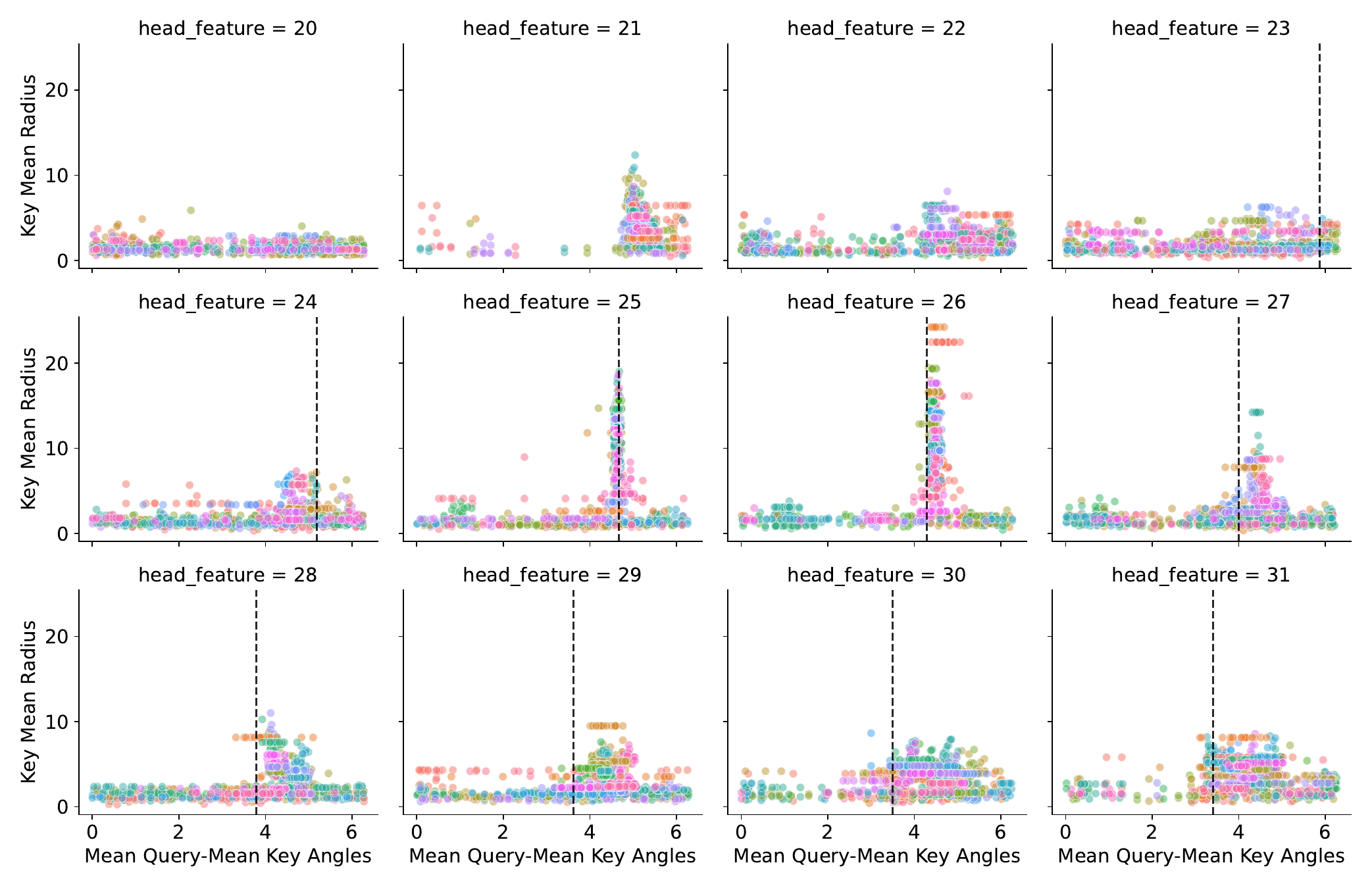}
    \caption{DeepSeek-V2-Lite: Mean radius of keys plotted against the angle between mean query and mean key for each layer (colored) and head for the last 12 rotary features. Black dashed lines for feature 23-31 are the lower bounds for the angle $\phi_i$ for each feature to serve as a rotary offset feature. Any feature without a dashed line are outside the bounds. The rotational keys are repeated 16 times to match the queries, so there are 16 times as many datapoints as there are actual rotational keys. We do note a single feature, feature 21, which is a low-frequency cyclic rotary feature that has feature instances with large radius in some heads. It may be the result of context length extension (see Section \ref{sec:llama_3_series_and_context_length_extension} for a potential explanation).}
    \label{fig:deepseek_angular_diff_vs_mean_radius}
\end{figure}

\section{Comparing Query-Key distributions for Wikipedia and Code Sample}
\begin{figure}[h]
    \centering
    \includegraphics[width=0.7\textwidth]{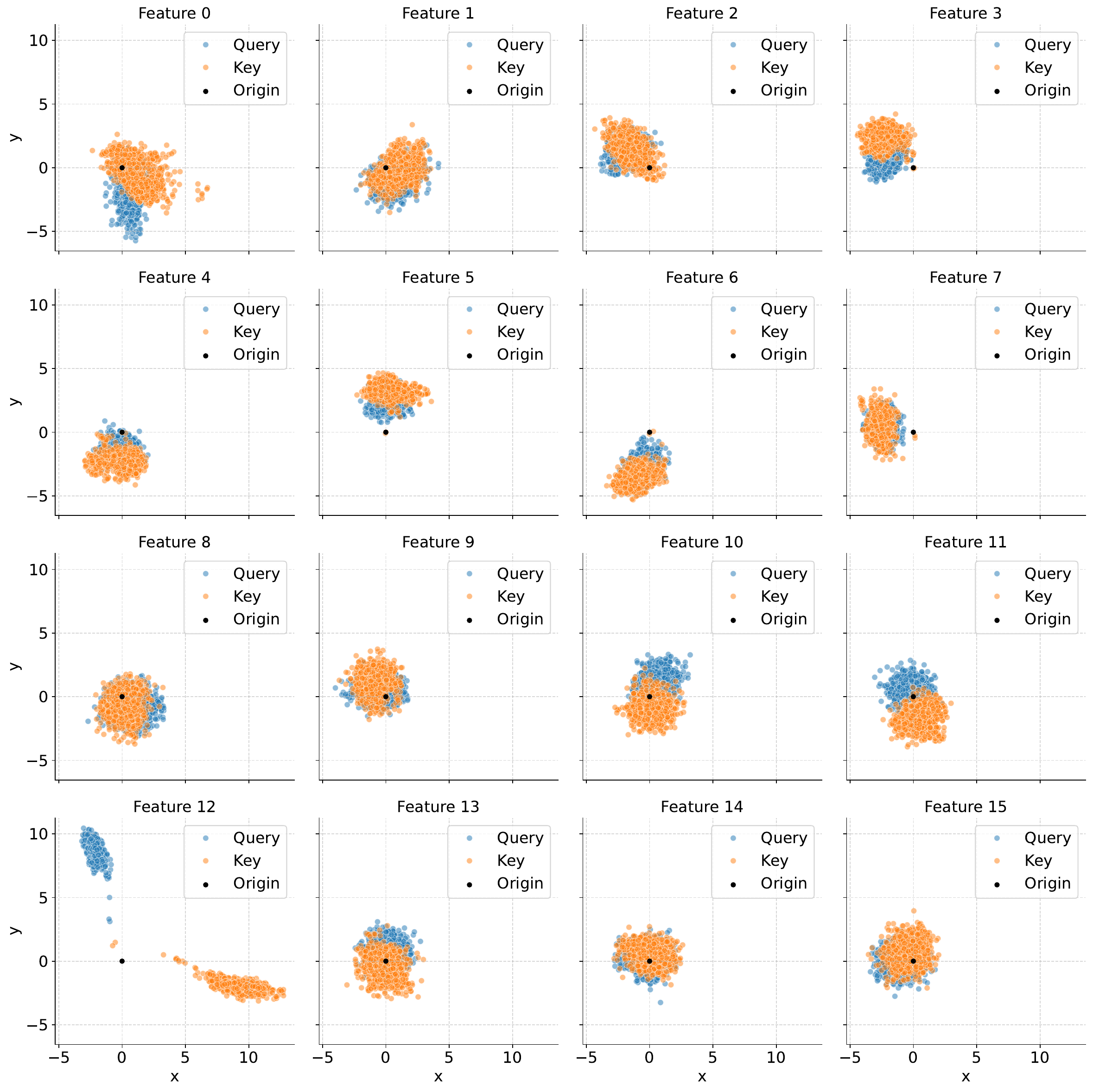}
    \caption{The x and y coordinates of the query and key vectors for features 0-15 in Phi-1 in the same head analyzed in the main text.}
    \label{fig:feature_12_xy_coords_wiki}
\end{figure}

\begin{figure}[h]
    \centering
    \includegraphics[width=0.7\textwidth]{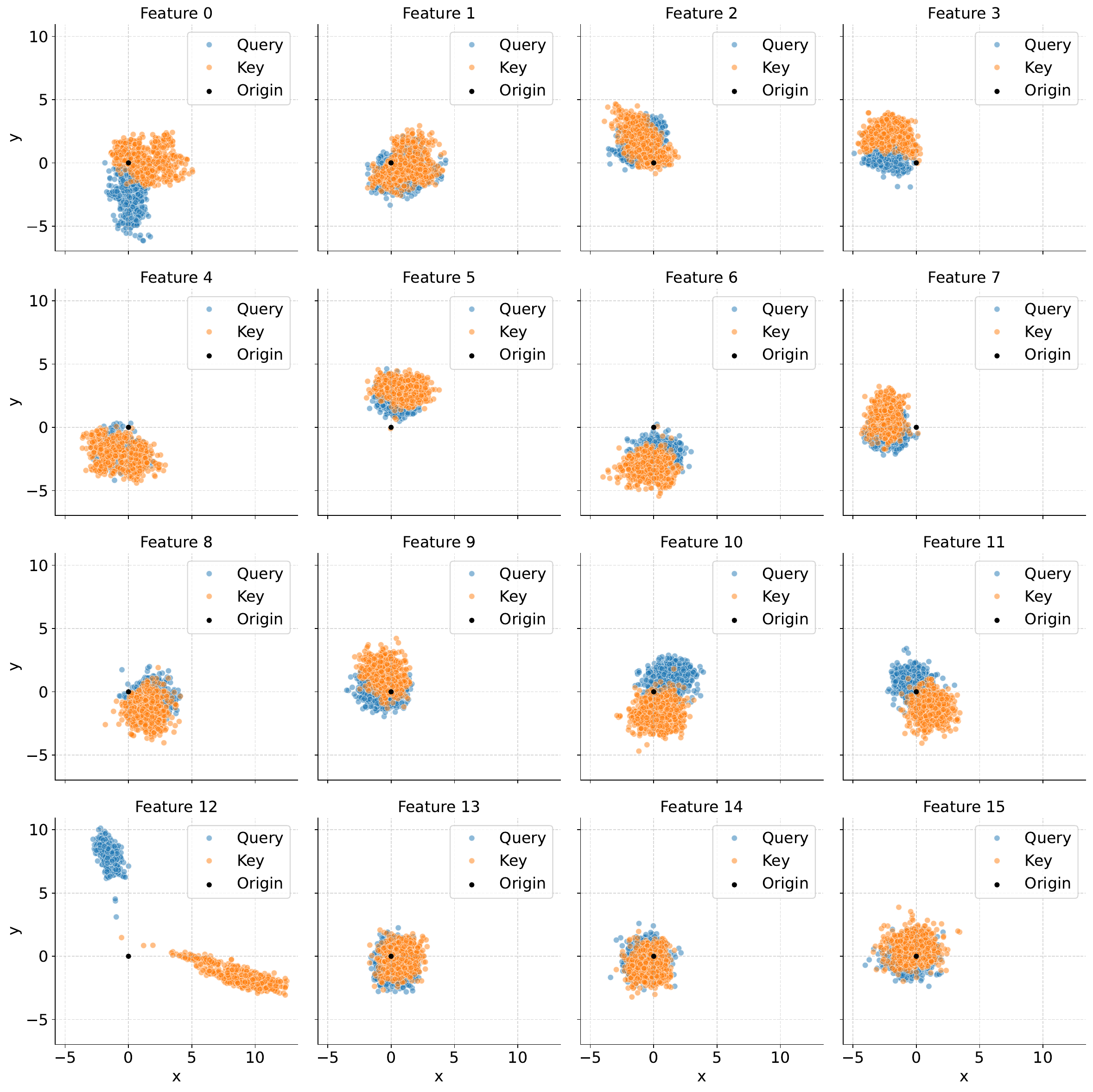}
    \caption{The x and y coordinates of the query and key vectors for features 0-15 in Phi-1 in the same head analyzed in the main text. But using the code sample instead of the wiki sample. Compare against Figure \ref{fig:feature_12_xy_coords_wiki}.}
    \label{fig:feature_12_xy_coords_code}
\end{figure}

\begin{figure}[h]
    \centering
    \includegraphics[width=0.7\textwidth]{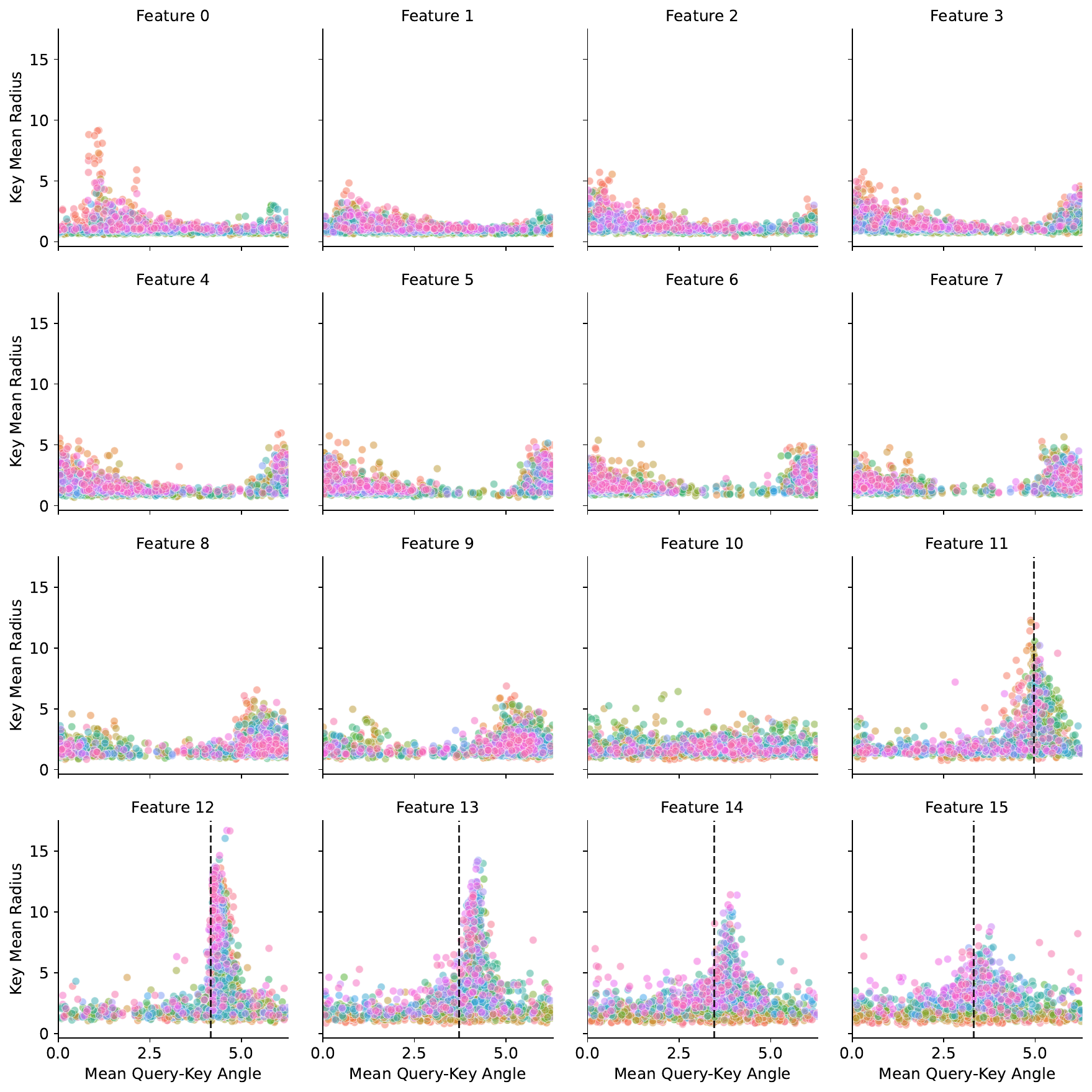}
    \caption{Mean radius of keys plotted against the angle between mean query and mean key for each layer (colored) and head for the last 12 rotary features using the code sample for Phi-1. Same plot as in Figure \ref{fig:angular_diff_vs_radius} but using the code sample for collecting activations.}
    \label{fig:angular_diff_vs_mean_radius_key_code}
\end{figure}

The analysis in the main text uses a data sample from Wikipedia, however the properties hold for many other samples. In Figures \ref{fig:feature_12_xy_coords_wiki} and \ref{fig:feature_12_xy_coords_code} we collect the activations for the same data as the main text and also for a code sample and plot the queries and keys for the rotary dimensions of Phi-1. Here we can clearly see that Feature 12 maintains the distributions and the angle of the queries and the keys between two different data samples. In Figure \ref{fig:angular_diff_vs_mean_radius_key_code} we reproduce the same plot as in Figure \ref{fig:angular_diff_vs_radius} but for the code sample to show that the properties hold for other data samples.

\section{Llama-3 series and context length extension}\label{sec:llama_3_series_and_context_length_extension}
\begin{figure}[h]
    \centering
    \includegraphics[width=0.7\textwidth]{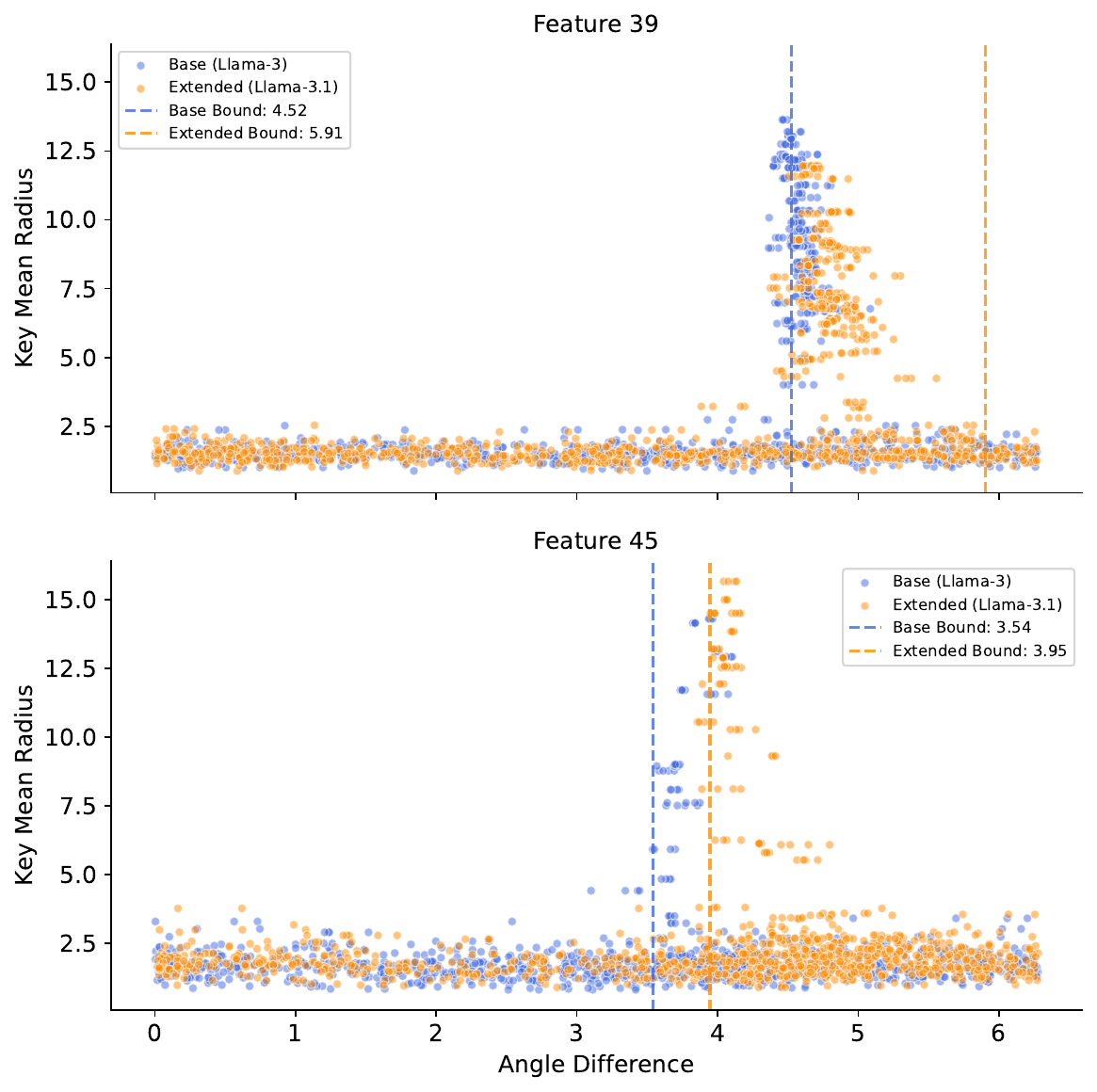}
    \caption{A comparison of the key mean radius and angle differences between the base and extended context length versions of Llama 8B 3 and 3.1 for features 39 and 45. In both cases, it can be observed that fine-tuning for extension moves the features closer to the extended bounds. Features with greater margin after extension (45) are seen to increase in radius and features with less margin (39) are seen to decrease in radius.}
    \label{fig:llama_angular_diff_vs_radius}
\end{figure}
\begin{figure}[h]
    \centering
    \includegraphics[width=0.7\textwidth]{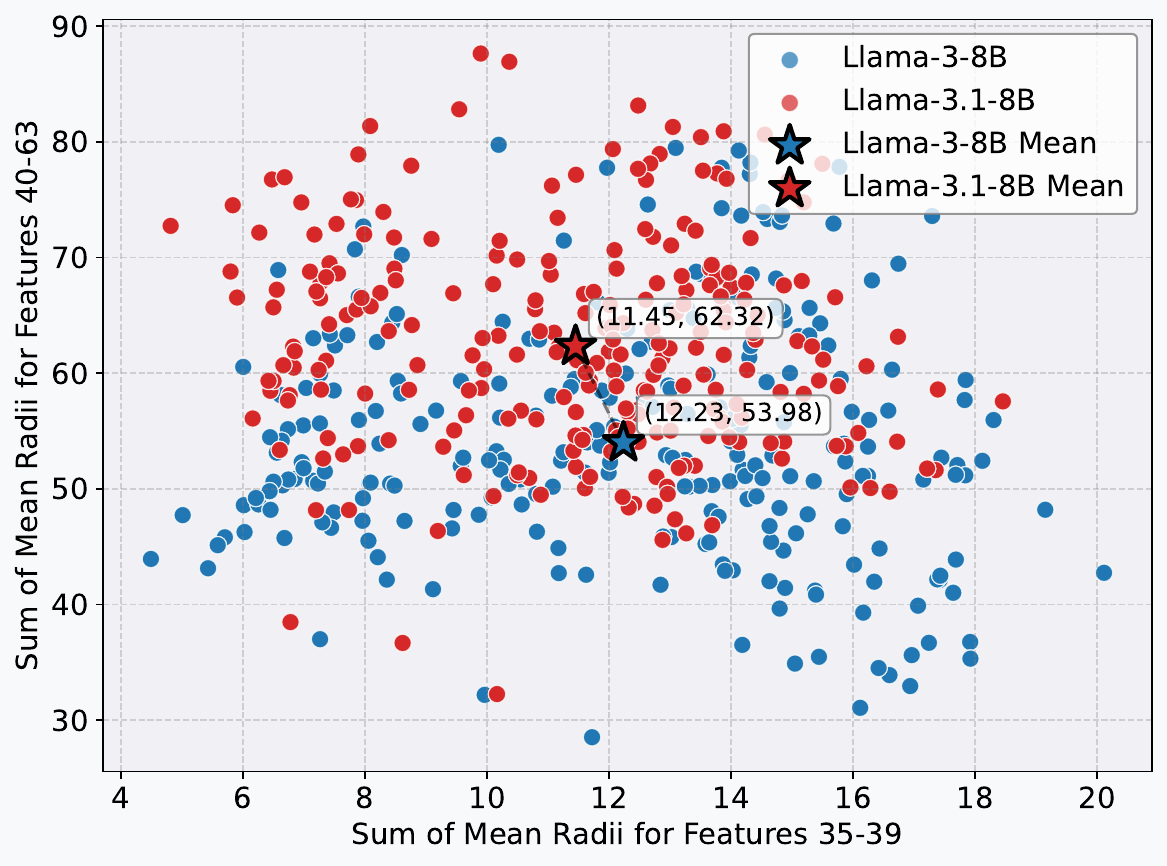}
    \caption{A comparison of the key mean radius for features that are no longer candidates for being rotary offset features after extension (features 35-39) and  features that are candidates for being rotary offset features both before and after extension (features 40-63) for every head in the models. With extension fine-tuning the sum of the radii decreases for features 35-39 and increases for 40-63.}
    \label{fig:change_in_radius_after_extension}
\end{figure}

We compare the basic and extended context length versions of Llama 8B 3 and 3.1 respectively. We firstly find that the basic version follows our bounds. The extended version, at first glance, looks to largely follow the base model's bounds instead of the extended bounds. Looking closer, we find that the features that after the extension have become outside or with low margin (close to $2\pi$) to our upper bound are decreasing in radius while the features that have an upper bound with larger margin are increasing in radius. The extended version's angle differences are also shifted closer to the extended bounds. Figure \ref{fig:llama_angular_diff_vs_radius} shows the key mean radius and angle differences between the base and extended context length versions of Llama 8B 3 and 3.1 for features 39 and 45. Figure \ref{fig:change_in_radius_after_extension} shows that the pattern seen in Figure \ref{fig:llama_angular_diff_vs_radius} with decreasing radii for features 35-39 and increasing radii for 40-63 is a pattern that holds across many heads.

In summary, some context length extension methods produce models that may not follow their new bounds, but instead follow something inbetween the base and extended bounds. The evolution of the radii and angular differences suggest the context length extension fine-tuning process may be adjusting the model towards the new extended bounds, but it is unclear whether it would converge with enough data. DeepSeek-V2-Lite is also a context length extended (which uses YaRN extension \citep{peng2023yarnefficientcontextwindow}) model, where our bounds fit well with the exception of feature 21. It is unclear whether feature 21 is a result of their context length extension method.

\begin{table}[h!]
    \centering
    \begin{tabular}{@{}lccccc@{}}
    \hline
    \textbf{Model} & \textbf{d\textsubscript{h}} & \textbf{r} & \textbf{p\textsubscript{max}} & \textbf{extended}\\
    \hline
    Llama-3-8B & 128 & 128 & 4096 & No\\
    Llama-3.1-8B & 128 & 128 & 131072 & Yes\\
    \hline
    \end{tabular}
    \caption{Model architecture details showing head dimension ($d_h$), rotary dimension ($r$), context length ($p_{max}$), and whether the linear projections for keys and queries have bias and whether the model was extended for longer context length.}
    \label{tab:llama_model_details}
\end{table}

A potential future research direction for making LLMs adopt context length extension more effectively could be to alter the distribution of rotary frequencies. The altered distribution should make sure all the rotary frequencies that would be effective as rotary offset features at the original context length are also effective at the extended context length. This could be accomplished by, for instance, having high frequency rotary features replace the first few rotary offset features at the base context length (thus making them invalid as rotary offset features at both context lengths) up to the point where the extended version of the upper bound starts.

\section{Data used for collecting activations}
We use the following text as input for collecting activation data from the models. It is an excerpt from a Wikipedia article on the United States of America. It is nearly 888 tokens if tokenized by Llama-3-8B tokenizer.
\begin{quote}
\small
The United States of America (USA), also known as the United States (U.S.) or America, is a country primarily located in North America. It is a federal republic of 50 states and Washington, D.C. as its federal capital district. The 48 contiguous states border Canada to the north and Mexico to the south, with the semi-exclavic state of Alaska in the northwest and the archipelagic state of Hawaii in the Pacific Ocean. The U.S. also asserts federal sovereignty over five major island territories and various uninhabited islands.[k] In addition, some 326 Indian reservations are treated as "domestic dependent nations" with tribal sovereignty rights. The U.S. is a megadiverse country, with the world's third-largest land area[d] and third-largest population, exceeding 340 million.[l] Its three largest metropolitan areas are New York, Los Angeles, and Chicago, and its three most populous states are California, Texas, and Florida.

Paleo-Indians migrated to North America across the Bering land bridge more than 12,000 years ago, and formed various civilizations and societies. Spanish exploration and colonization led to the establishment in 1513 of Spanish Florida, the first European colony in what is now the continental United States. France also began to colonize at this time, but major settlements came much later. Subsequent British colonization led to the first settlement of the Thirteen Colonies in Virginia in 1607. Intensive agriculture in the rapidly expanding Southern Colonies encouraged the forced migration of enslaved Africans. Clashes with the British Crown over taxation and political representation sparked the American Revolution, with the Second Continental Congress formally declaring independence on July 4, 1776.

The United States emerged victorious from the 1775–1783 Revolutionary War and expanded westward across North America, dispossessing Native Americans as it fought the Indian Wars. Expansion continued when the U.S. signed the 1803 Louisiana Purchase with Napoleonic France and won the Mexican–American War in 1848. As more states were admitted, a North–South division over slavery led to the secession of the Confederate States of America, which fought the Union in the 1861–1865 American Civil War. With the victory and preservation of the United States, slavery was abolished nationally. By the late 19th century, the United States established itself as a great power with victory in the Spanish–American War, a status solidified with its participation in World War I. Following Japan's attack on Pearl Harbor in December 1941, the U.S. entered World War II. The aftermath of the war left the U.S. and the Soviet Union as the world's two superpowers which led to the Cold War, during which both countries struggled for ideological dominance and international influence. The end of the Cold War and the Soviet Union's collapse in 1991 left the U.S. as the world's sole superpower, with significant geopolitical influence globally.

The U.S. national government is a presidential constitutional federal republic and liberal democracy with three separate branches: legislative, executive, and judicial. It has a bicameral national legislature composed of the House of Representatives, a lower house based on population, and the Senate, an upper house based on equal representation for each state. The country's Democratic and Republican parties have dominated American politics since the 1850s. Federalism provides substantial autonomy to the 50 states, while American values are based on a political tradition that draws its inspiration from the European Enlightenment movement. A melting pot of many ethnicities and customs, the culture of the United States has been shaped by centuries of immigration, and its soft power influence has a global reach.

One of the world's most developed countries, the U.S. ranks among the highest in economic competitiveness, productivity, innovation, human rights, and higher education. The United States accounted for over a quarter of nominal global economic output in 2024, and its economy has been the world's largest by nominal GDP since about 1890. It possesses by far the largest amount of wealth of any country and has the highest disposable household income per capita among OECD countries, though U.S. wealth inequality is higher than in most other developed countries. The U.S. is a member of multiple international organizations and plays a leading role in global political, cultural, economic, and military affairs.
\end{quote}
\textit{Source: Wikipedia article on the United States, accessed March 2025}

We use the following text as input for collecting activation data from the models for the Code Sample (used for Figure \ref{fig:feature_12_xy_coords_code}). It is code generated by GPT-5.
\small
\begin{verbatim}
"""
Compact Python utility for parallel URL fetching with optional file caching.
"""
from __future__ import annotations
import hashlib, sys, threading, urllib.request
from concurrent.futures import ThreadPoolExecutor, as_completed
from dataclasses import dataclass
from datetime import datetime, timezone
from pathlib import Path
from typing import List, Optional

@dataclass
class FetchResult:
    url: str; status: int; content: bytes; fetched_at: datetime

class FileCache:
    def __init__(self, directory: Path):
        self.dir = directory; self.dir.mkdir(parents=True, exist_ok=True)
    def _path(self, key: str) -> Path:
        return self.dir / f"{hashlib.sha256(key.encode()).hexdigest()}.cache"
    def get(self, key: str) -> Optional[bytes]:
        p = self._path(key); return p.read_bytes() if p.exists() else None
    def set(self, key: str, data: bytes) -> None:
        p = self._path(key); p.write_bytes(data)

def fetch_url(url: str, timeout: int = 10) -> FetchResult:
    req = urllib.request.Request(url, headers={"User-Agent": "pyfetch/1.0"})
    try:
        with urllib.request.urlopen(req, timeout=timeout) as r:
            return FetchResult(url, getattr(r, "status", 200), r.read(), datetime.now(timezone.utc))
    except Exception as e:
        return FetchResult(url, 599, str(e).encode(), datetime.now(timezone.utc))

def parallel_fetch(urls: List[str], workers: int = 6, cache: Optional[FileCache] = None) -> List[FetchResult]:
    results = []
    with ThreadPoolExecutor(max_workers=workers) as ex:
        futures = {ex.submit(fetch_url, u): u for u in urls if not (cache and cache.get(u) and results.append(FetchResult(u,200,cache.get(u),datetime.now(timezone.utc))))}
        for fut in as_completed(futures):
            res = fut.result(); results.append(res)
            if cache and res.status == 200: cache.set(res.url, res.content)
    return results

def main(argv=None) -> int:
    argv = argv or sys.argv[1:]
    if not argv: return 2
    cache_dir, urls = None, []
    if "--cache" in argv:
        idx = argv.index("--cache"); cache_dir = Path(argv[idx+1]); del argv[idx:idx+2]
    urls = argv; cache = FileCache(cache_dir) if cache_dir else None
    res = parallel_fetch(urls, cache=cache)
    for r in res:
        print(r.url, r.status, len(r.content))
    return 0

if __name__ == "__main__":
    raise SystemExit(main())
\end{verbatim}
\textit{Source: GPT-5 generated code sample, accessed August 2025}
%%%%%%%%%%%%%%%%%%%%%%%%%%%%%%%%%%%%%%%%%%%%%%%%%%%%%%%%%%%%%%%%%%%%%%%%%%%%%%%
%%%%%%%%%%%%%%%%%%%%%%%%%%%%%%%%%%%%%%%%%%%%%%%%%%%%%%%%%%%%%%%%%%%%%%%%%%%%%%%
\end{document}